\documentclass{article}
\usepackage[ruled]{algorithm}
\usepackage{algpseudocode}
\usepackage{graphicx}
\usepackage{subcaption}
\usepackage{amsmath}
\usepackage{amsfonts}
\usepackage{amssymb}
\usepackage{amsthm}
\usepackage{mathtools}
\usepackage{booktabs}
\usepackage{wrapfig}
\usepackage{hyperref}
\usepackage{tikz}
\usetikzlibrary{positioning, arrows.meta, shapes}

\title{Neural Human Pose Prior}

\author{ Michal Heker\\Yoom\\michal@yoom.com \and Sefy Kagarlitsky\\Yoom\\ sefy@yoom.com \and David Tolpin\\Yoom\\davidt@yoom.com}

\begin{document}

\maketitle
\begin{abstract}
We introduce a principled, data-driven approach for modeling a
neural prior over human body poses using normalizing flows.
Unlike heuristic or low-expressivity alternatives, our method
leverages RealNVP to learn a flexible density over poses
represented in the 6D rotation format. We address the challenge
of modeling distributions on the manifold of valid 6D rotations
by inverting the Gram-Schmidt process during training, enabling
stable learning while preserving downstream compatibility with
rotation-based frameworks. Our architecture and training
pipeline are framework-agnostic and easily reproducible. We
demonstrate the effectiveness of the learned prior through both
qualitative and quantitative evaluations, and we analyze its
impact via ablation studies. This work provides a sound
probabilistic foundation for integrating pose priors into human
motion capture and reconstruction pipelines.
\end{abstract}

\section{Introduction}
\label{sec:introduction}

Human motion capture is concerned with obtaining a parametric
representation of human motion from raw inputs such as
single-view or multiview videos. Even in a studio setup, the
inputs are often noisy and partial --- consider, for example, a
group of people moving together, casting shadows and occluding each
other relative to the camera. A common consensus is that some
population-based bias --- a prior --- should be introduced into
the model for more reliable motion capture.  However,
constructing or learning a human \textit{motion} prior is too
challenging due to both scarcity of data and complexity and
diversity of human motions. A viable compromise though is to 
come up with a human \textit{pose} prior. A parametric human
body model commonly has only a few dozen joints with at most
three degrees of rotational freedom each, a dimensionality well
within the capabilities of modern machine learning approaches. 

Intuitively, a prior should be a density function over human pose
parameter vectors such that the gradient of the function points
in the direction of the quickest `improvement' of the pose. In
addition, since human poses lie on a
low-dimensional manifold embedded into the Euclidean space of
pose parameters, the ambient pose space should be projected into 
a latent space in which sufficiently small Euclidean
vicinity  of a viable pose consists of poses that are only
slightly more or less viable. These two constituents ---
reparameterization and a density function --- are the essentials
of a body pose prior. 

Despite the apparent simplicity of the task, it turns out that
coming up with a coherent, scientifically justified procedure of
constructing a human body prior is quite challenging and hard to
get right (Section~\ref{sec:related}). In this work, we
establish a clean, principled scheme for training and using
during inference a human pose prior. Our contributions are as
follows:

\begin{itemize}
\item  We establish a systematic foundation of a data-driven
human pose prior.

\item We describe, in a framework-agnostic way and in detail sufficient
for easy reproduction, our network architecture and training
routine.

\item  We evaluate our approach empirically, quantitatively and
qualitatively, compare with baselines, and support our design
choices by ablation studies.
\end{itemize}

\section{Background}
\label{sec:background}

\subsection{Losses vs. Probabilities}

\subsubsection{Losses}

Consider a task of finding a number the square of which is 9.

$$x^2 = 9$$

An obvious solution is the closed-form one. If $f(x) = x^2$ then
$f^{-1}(y) = \sqrt{y}$, so if $x^2 = 9$ then $x = \sqrt{9} = 3$.
However, we do not always have $f^{-1}$ in closed form. An
alternative is to use a numerical method of sort, for example,
gradient descent:

\begin{itemize}
\item Parameter: $x$
\item Ground truth: $9$
\item Transformation: $x^2$
\item Loss ($L_2$): $(9-x^2)^2$
\end{itemize}

$$\nabla_x (9-x^2)^2 = -4(9-x^2)x$$

If we make a small step in the direction of $4(9-x^2)x$, usually
proportional to the gradient, we will hopefully get closer to
the solution. 

\begin{enumerate}
\item Initially: $x \gets 2$
\item Learning rate: $0.025$
\item Gradient $-4(9-4)2 = -40$
\item $x \gets x + 40 * 0.025 = 3$
\item New gradient: $-4(9-9)3 = 0$ --- voila!
\end{enumerate}

\subsubsection{Unidentifiable Problems}

Sometimes, however, many solutions are possible; consider

$$x\cdot y = 9$$

There are infinitely many solutions, to list just some of them:
$x=1, y=9$, $x=3, y=3$ $x=900, y=0.01$. If we run gradient
descent, then, depending on the starting point (and, in general,
other random factors) we are going to get different solutions.
How can we make the result deterministic? 

We should decide on what results are more desirable. For
example, we may want a result as close to the straight line
$x-y=0$ as possible.  Then, our loss will combine two terms:

\begin{enumerate}
\item What assignment constitutes a solution? In our case a perfect
solution is when $9-x\cdot y=0$, so the loss, let us call it
$L_c$, is $L_c(x, y) = (9 - x\cdot y)^2$. 
\item What kind of solutions do we prefer?  We want $x$ and
$y$ to be close to each other, hence our preference loss, $L_p$,
is $L_p(x, y) = (x - y)^2$. 
\end{enumerate}

Note here that our preference loss is quite general; we can
apply it to many problems involving an equation connecting
between $x$ an $y$. Our total loss $L$ is then 

$$L=L_p + L_c = (x-y)^2 + (9 - x\cdot y)^2$$

and its gradient is

\begin{align*}
\nabla_x L & = 2(x-y) - 2(9 - x \cdot y)y \\
\nabla_y L & = -2(x-y) - 2(9 - x \cdot y)x
\end{align*}

Gradient descent, with sensible parameters, will bring us to
$x=3, y=3$ or $x=-3, y=-3$, the kind of solution we have been
looking for.

\subsubsection{Bayesian Statistics}

Let us now connect our intuition about two kinds of losses to
the fundamentals of machine learning --- the Bayesian statistics. In
Bayesian statistics, a model is defined with respect to a latent
variable $x$, which is probabilistically connected to an
observed variable $y$. A Bayesian model is generally written
as follows:

\begin{align*}
x & \sim \mathrm{Prior} \\
y & \sim \mathrm{Conditional}(x)
\end{align*}

The first line says that $x$ is distributed according to
distribution $\mathrm{Prior}$, and the second line says that $y$
is, conditionally on $x$ (that is, for any given value of $x$),
is distributed as $\mathrm{Conditional}(x)$. 

The most important manipulation we perform on the model is
called ‘posterior inference’. An observation $y=Y$ is given, and
we want to know how $x$ is distributed given that observation.
To answer this question, we observe that when a random variable is
drawn from a distribution, every assignment of the random
variable appears with a certain \textit{probability}. We write
$p_{\mathrm{Prior}}(x)$ or just $p(x)$ for the probability of
drawing some value $x$ from $\mathrm{Prior}$, and
$p_{\mathrm{Conditional}}(y|x)$ or just $p(y|x)$ for the
probability of drawing some value $y$ from
$\mathrm{Conditional}$ given a certain value of $x$.
When we say that we want to know the posterior distribution of
$x$ given $y=Y$, this means that we want to be able to compute,
exactly or approximately, $p(x|y=Y)$, that is, the probability
of $x$ given that we know that $y=Y$ was drawn from
$\mathrm{Conditional}(x)$.

The Bayes rule, which is a simple consequence of the definitions
of probabilities, states that 

$$p(x|y=Y) = \frac {p(x) p(y=Y|x)} {p(y=Y)} $$

The Bayes rule takes us halfway towards the answer, but we
still do not know $p(y=Y)$ (which is called ‘marginal evidence’).
Fortunately, since for a given evidence $p(y=Y)$ is a constant,
there are ways to obtain $p(x|y=Y)$ without knowing $p(y=Y)$, and
we just write:

$$p(x|y=Y) \propto p(x) p(y=Y|x)$$

The equation above is the most important formula in Bayesian
statistics and serves as the basis for machine learning. In
machine learning, we are often interested in the most probable
assignment to $x$, which we can find by maximizing the right-hand
side, e.g. by gradient ascent.

How is this all related to losses? Most computations on
probabilities are performed in log space, that is, on the
logarithms of probabilities. There are two reasons for this.

\begin{enumerate}
\item Probabilities are constrained between 0 and 1 for probability
masses and between 0 and some positive number for
probability densities. $\log$ expands the axis,  making the
computations more accurate.
\item In the formulae above, the probabilities are multiplied
and divided. Multiplication  and division are cumbersome
operations, prone to precision loss and overflows/underflows. It
is safer to add and subtract, and multiplications and divisions
in natural space correspond to additions and subtractions in log
space.
\end{enumerate}

We can rewrite our Bayes rule equation in log space:

$$\log p(x|y=X) = \log p(x) + \log p(y=Y|x) + C$$

Losses go up when probabilities go down, quite intuitively.
The first term on the left-hand side is just our $L_p$, the
preference (or the prior) loss, negated:

$$L_p \equiv - \log p(x)$$

Similarly, the second term is our negated conditional, or data, loss:

$$L_c \equiv - \log p(y=Y|x)$$.

We are familiar with the need for the data loss, and now the
reason behind the prior loss is clear as well --- without the
prior loss we cannot compute the posterior probability, or find
the maximum \textit{a posteriori assignment} to $x$! Sometimes,
models are formulated without $p(x)$, or $L_p$, explicitly
written down, but that does not mean there is no $p(x)$, it just
means that $p(x)$ is ‘non-informative’, that is, we agree to
accept any solution that fits the data. In most cases, however,
the prior loss is important.

\subsection{Constructing a Prior}

We know by now that a prior loss is the negative log probability
function of some distribution. What distribution is suitable as
the prior distribution for a particular problem?
One way is just to pull our socks up and construct a prior
distribution based on our knowledge of the problem domain, e.g.
of the human body and rigid mechanics. We know what reasonable
angles of knees, elbows, spine vertebrae, etc are. We can, with
some effort, put reasonable losses on each joint separately,
constrain movements, and then combine them to get the pose
joint distribution. However, this approach is laborious and
error-prone. It is also quite hard to debug: we won't know that
our prior is not good until we stumble on an inferior
optimization outcome and have to tweak our hand-crafted prior
again to fix the flaw (and probably introduce another problem).

Another way is to somehow learn the prior from data. Learning
priors from data is not novel, it is known for half a century as
‘empirical Bayes’. Using a large dataset of all possible poses,
we can estimate parameters of a distribution that supposedly
fits the data well enough. The problem, however, is again how to
choose such a distribution. There are some parametric options,
such as Gaussian mixture models. However, Gaussian mixture models
are not expressive enough for high-dimensional manifolds, and a
human body model is at least a 66-dimensional distribution,
which is too tough.

\subsubsection{Neural Density Estimation}

A modern approach to constructing distributions from data is
\textit{neural density estimation}~\cite{PNR+21}. For neural
density estimation, we need a neural architecture which:

\begin{itemize}
\item Represents a distribution that fits well the dataset.
\item Supports efficient sampling from the distribution.
\item Supports computing the probability of a sample from the
distribution.
\end{itemize}

We only need the third point to use the model to compute the
prior loss. But to achieve the 3rd point, we also need to ensure
that the 1st point holds. The 2nd point helps model validation
even if we do not intend to sample from the prior.

\subsubsection{Normalizing Flows}

An overall idea of a deep model for density estimation (learning
a distribution from a dataset) is based on transforming a sample
from a known distribution to a sample from the data
distribution, or vice versa. The forward transformation serves
sampling, the backward transformation serves density
computation. So far, the big picture is quite straightforward:

\begin{figure}[H]
\centering
\begin{tikzpicture}
\tikzset {
	module/.style = {
			draw,rectangle,thick,
			minimum width=2em,minimum height=1.7em,
	},
	arr/.style = {Stealth-Stealth,thick},
}
\node[module] (draw) {draw $z$ from $\mathcal{N}(0, \mathbb{I})$};
\node[module,right=of draw] (transform) {transform via NN};
\node[module,right=of transform] (x) {x};
\draw[arr] (draw) -- (transform);
\draw[arr] (transform) -- (x);
\end{tikzpicture}
\end{figure}

Provided we managed to train the NN at the second step, the
forward pass is straightforward:

\begin{align}
z & \sim \mathcal{N}(0, \mathbb{I}) \\ \nonumber
x & = T(z)
\label{eqn:flow}
\end{align}

The difficulty arises in the backward pass, for density
computation. Naively, one would assume that all that is needed
is to invert $T$ (which is not trivial, but can be accomplished)
and compute $p(T^{-1}(x);\,\mathcal{N}(0, \mathbb{I}))$. However,
this computation does not take into consideration the change in the
differential volume under $T$ --- an infinitely small hypercube in
the target $x$ space does not have the same volume as the
corresponding infinitely small ‘hyperbrick’ in the latent $z$
space. The correct formula is obtained by multiplying the latent
space density by the determinant of the Jacobian of the inverse
transformation.

\begin{equation}
p(x) = p(T^{-1}(x);\,\mathcal{N}(0, \mathbb{I}))\cdot\left|\det \mathbb{J}(T^{-1}(x))\right|
\label{eqn:p_x}
\end{equation}

Equation~\eqref{eqn:p_x} means three things:

\begin{enumerate}
\item The cardinality of $x$ and the cardinality of $z$ must be
the same, otherwise the Jacobian does not have a determinant.
\item The transformation must be non-degenerate everywhere, the
determinant of the Jacobian cannot be zero for any value in the domain,
meaning it is either everywhere positive or everywhere negative.
\item The NN representation of $T$ must allow an efficient
computation of the Jacobian. In principle, automatic
differentiation can be employed, in practice this would mean
nested automatic differentiation (because the density computation
itself must pass through autodiff for training), rendering
implementation impractical.
\end{enumerate}

A network architecture possessing such properties is called a
\textit{normalizing flow}. There are many variants of
\textit{flows} in use for density estimation, which differ in
their expressivity, sampling and density computation
performance. In the next section we discuss one particular architecture,
RealNVP, which strikes a compromise between expressivity,
performance, and ease of implementation. This is the
architecture we use in our implementation of the pose prior.

\subsubsection{RealNVP}

\begin{figure}
\begin{tikzpicture}
\tikzset{
  arr/.style = {-Stealth, thick},
  tensor/.style = {draw, diamond, align=center},
  op/.style = {draw, circle, align=center},
}

\node[tensor] (x) {$x$};
\node[op, right=of x] (split) {$\prec$};

\node[tensor, above right=of split] (x1)  {$x_1$};
\node[tensor, below right=of split] (x2)  {$x_2$};

\node[op, right=of x2] (mul) {$\times$};
\node[op, right=of mul] (add) {$+$};
\node[op, above=of mul, xshift=8pt] (t) {$t$};
\node[op, left=of t, xshift=2pt] (s) {$s$};
\node[op, above right=of t, yshift=-2pt] (id) {$=$};
\node[tensor, right=of id] (y2) {$x'_2$};
\node[tensor, right=of add] (y1) {$x'_1$};
\node[op, below right=of y2] (cat) {$\succ$};
\node[tensor, right=of cat] (y) {$x'$};

\draw[arr] (x) -- (split);
\draw[arr] (split) -- (x1);
\draw[arr] (split) -- (x2);
\draw[arr] (x2) -- (mul);
\draw[arr] (mul) -- (add);
\draw[arr] (add) -- (y2);
\draw[arr] (x1) -- (s);
\draw[arr] (x1) -- (t);
\draw[arr] (s) -- (mul);
\draw[arr] (t) -- (add);
\draw[arr] (x1) -- (id);
\draw[arr] (id) -- (y1);
\draw[arr] (y1) -- (cat);
\draw[arr] (y2) -- (cat);
\draw[arr] (cat) -- (y);

\end{tikzpicture}
\caption{A RealNVP layer}
\label{fig:realnvp}
\end{figure}
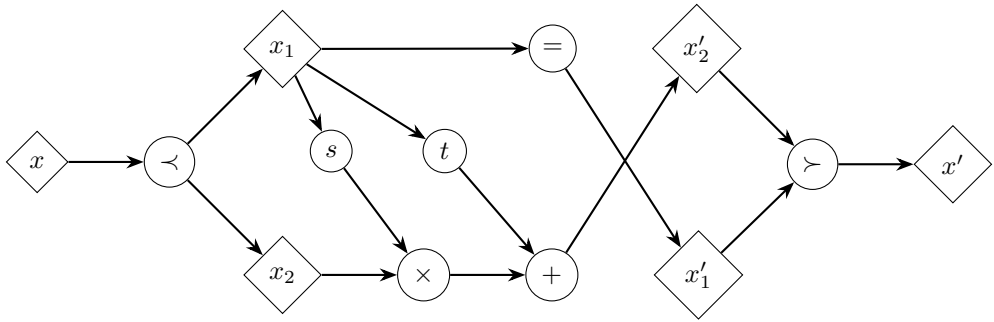

RealNVP\cite{DSB17} is a normalizing flow that uses a clever
trick to ensure that the transformation is invertible and the
Jacobian is easily computable. It is built of a stack of
\textit{affine coupling} layers (Figure~\ref{fig:realnvp}).
Every layer is just an affine (linear) transformation, applied
to a part (usually a half) of the inputs, while the rest of the
inputs are used to compute, through another ‘hyper’-network, the
coefficients of the transformation. The inputs that are
transformed and those used to compute the transformation are
swapped before each layer.

We will not get into mathematical details of RealNVP here. The
cited paper provides a clear and detailed explanation of its
inner working. What is important to know though is that it is a
relatively big model (the affine backbone, with a separate
hypernet attached to each layer of the backbone), which, when
trained, looks like a regular distribution object. It exposes
two methods: 

\begin{itemize}
\item $\mathtt{sample}$ that draws a sample from the learned
distribution; and
\item $\mathtt{logprob}$ that computes the log probability
density of a sample.
\end{itemize}

For a given data entry $\mathtt{x}$ the prior loss thus can be
computed
as $-\mathtt{logprob(x)}$.

\section{Related Work}
\label{sec:related}

Research on human pose priors falls into one of two categories
--- works which formulate construction of a prior as a density
estimation task~\cite{PNR+21,KPJ+21,ZBX+20,BAL+16} and works
that consider heuristic approximation which is not based on
probabilities~\cite{HTB+24,CMW+23,PCG+19}. Our contribution
belongs to the former category.

In the density estimation category, an early work~\cite{BAL+16}
employed a Gaussian mixture model prior, a theoretically sound
approach which, unfortunately, is limited in expressivity and
cannot represent the rich structure of the high-dimensional pose
distribution. With the raise of neural density estimation with
normalizing flows~\cite{PNR+21}, several works employed variants
of normalizing flows to train human pose priors, in 
different contexts. \cite{KPJ+21,ZBX+20} learn a pose prior and
propose to optimize in the latent space. However, both works
use simple losses (quadratic penalties) in the latent space
for optimization under the assumption that the Jacobian
determinant does not depend on the latent variable.
However, RealNVP and similar flows are not volume-preserving:
its affine coupling layers introduce input-dependent scaling,
making the transformation from latent to data space nonlinear
and volume-changing. Consequently, losses defined in latent
space do not correspond to the same geometry in ambient space, and such optimization leads to unintended distortions. In
addition, \cite{KPJ+21} struggles with the 6D rotation
representation~\cite{ZBL+19} in the ambient space. 6D rotations
do not induce a proper density because they are invariant to
scaling. \cite{DSP24} propose a flow architecture specifically
tailored to product spaces of SO(3) manifolds; while
theoretically solid and supported by empirical evidence, this
approaches involves a complicated implementation and imposes
restrictions on the ambient space. Our work introduces a
theoretically rigorous and practically robust approach to
training and inference with a neural pose prior with proper
training losses and principled handling of 6D rotation
representation of the ambient space.

In the heuristic category, the first practical architecture was
based on a variational autoencoder~\cite{PCG+19}. However, a
variational autoencoder is an architecture inappropriate for
density estimation: then density can only be computed
through importance sampling and even then is approximated by
the evidence lower bound. Optimizing in the latent space is not
sound either and practically produces inferior results, because
the latent space is only approximately, rather than exactly as
in the case of normalizing flows, a multi-variate normal
distribution. Recent works explored modeling the variety of
human poses using Riemannian distance fields~\cite{HTB+24}, GAN
networks with constrained latent spaces~\cite{DRC+22}, and
gradient fields of score-based generative models~\cite{CMW+23}.
While some of these works produce interesting empirical
evidence, they are not based on a solid foundation of
probabilistic reasoning.

\section{Dataset}
\label{sec:dataset}
Our proprietary motion dataset is the result of years of
dedicated effort in capturing and processing high-quality
volumetric video data using Yoom's volumetric system. Collected
in-house, it represents a diverse range of human motions
recorded in controlled environments. Each volumetric video
sequence is topologically aligned using a registration algorithm
that leverages both geometry and texture. Subsequently, the data
is processed by fitting a parametric human body model to the
footage using an optimization technique that, while
time-consuming, yields highly accurate representations of body
pose and shape.

This rigorous process has produced a dataset which is organized
as a list of motion sequences, each representing continuous
human motion of a subject. Every sequence includes a fixed body
shape represented by a vector and a series of frames. Each frame
contains joint rotations, representing the pose of the body, and
a global root translation, indicating the position of the body
in 3D space. For the purpose of training the human pose prior, we
use the non-root joint rotations of all poses in the dataset.

\section{Pose Prior on 6D Rotations}
\label{sec:degramschmidt}

6D representation of rotations~\cite{ZBL+19}
is arguably the most suitable representation for global
optimization of body poses, and it is the representation that we
use for many tasks. Therefore, the prior should be constructed
with respect to 6D rotations. However, 6D representation space is
both restricted and improper probabilistically: multiplying a 6D
vector by any positive number gives a 6D vector corresponding to
exactly the same rotation; hence the density of any particular
6D vector is zero. A seemingly reasonable workaround is to learn
a prior for the orthonormal 6D representation, where the two 3D
vectors composing the representations are orthogonal to each
other and with a magnitude of 1.  This workaround, however, has
its own drawbacks:

\begin{itemize}
\item The prior for the orthonormal 6D representation is a
3-dimensional manifold in 6D space, which is hard to learn, even
approximately, with flow models. The orthonormal 6D prior
concentrates in a small part of the space, leaving most of the
space empty.
\item The network is trained to project the orthonormal 6D
ambient space into the latent space. Projection in the opposite
direction, from the latent space to the ambient space, may not
always yield a point on the orthonormal 6D manifold, due to
inherent approximation of transformation by a neural network.
\end{itemize}

The training can, in principle, be regularized by imposing the
orthonormality loss on the outcome of projection from the latent
space to the ambient space. However, such regularization just
makes the problem less acute. A sound and robust solution is
to relax, for training, the orthonormality constraint by inverting
the Gram-Schmidt process applied to 6D vectors.

Formally, let each rotation be represented in 6D as two
orthonormal 3D vectors $\pmb{b}_1, \pmb{b}_2 \in
\mathbb{R}^3$, i.e.,

\begin{equation}
\pmb{b}_1, \pmb{b}_2 \in \mathbb{R}^3, \quad \text{with } \|\pmb{b}_1\| = \|\pmb{b}_2\| = 1, \quad \pmb{b}_1^\top \pmb{b}_2 = 0
\end{equation}

Non-orthonormal vectors $\pmb{a}_1,\,\pmb{a}_2 \in
\mathbb{R}^3$ representing the same rotation as $\pmb{b}_1,\,
\pmb{b}_2$ are randomly
constructed such that Gram-Schmidt recovers $\pmb{b}_1,\,
\pmb{b}_2$ from them. Let

\begin{equation}
\begin{aligned}
\rho_1, \rho_2 & \sim \text{Gamma}(k, \theta = 1/k) \\
\alpha & \sim \mathcal{N}(0, \sigma^2)
\end{aligned}
\end{equation}

Then 

\begin{equation}
\begin{aligned}
\pmb{a}_1 &= \rho_1 \cdot \pmb{b}_1 \\
\pmb{a}_2 &= \rho_2 \cdot \pmb{b}_2 + \alpha \cdot \pmb{a}_1
\end{aligned}
\end{equation}

This augmentation allows both efficient training and smooth
inference in either the latent or the ambient space. Parameters
$k$ and $\sigma$ control the level of deviation from
othronormality. The stronger the deviation is, the easier the
training, but the obtained prior becomes `flatter', which should
be accounted for during inference, e.g. by adjusting
the weight of the prior-based regularization loss.

\section{Network, Training, and Inference}
\label{sec:network}

\subsection{Network}

At the core of the model is a RealNVP instance. The modeled 
distribution's dimension is 126, corresponding to 21 non-root
joints and 6 components per joint. We do not include the root 
joint in the prior because the orientation of the root joint
reflects the studio and the camera setup rather than the body 
pose. There are 12 affine coupling layers. Each layer's hypernet
consists of 4 layers of 256 neurons each.

RealNVP works by swapping halves of the input/latent vector
on input to each layer, with one half being transformed by
the affine coupling layer and the other being used for 
computing the coefficients of the affine transformation.
Thus, it makes sense that each half of the input vector has
components related to the other half. To facilitate this, we
\textit{transpose} the pose tensor of dimension $21 \times 6$
into dimension $6 \times 21$ before flattening.  This places
the first row of 6D rotation of each joint into the first half of
the vector, and the second row into the second half of the
vector. During sampling, the pose vector can again be transposed
to the conventional form. 

\subsection{Training} 

The training is straightforward. RealNVP is trained with the
usual log-probability loss on the data using the Adam optimizer
with early stopping on a held-out validation set. Each training
batch passes through the inverse Gram-Schmidt process
(Section~\ref{sec:degramschmidt}) with $k=100$ and $\sigma=0.1$.

\subsection{Inference}

The inference can be performed either in the latent or in the
ambient space, depending on an application. When
optimizing in the ambient space, for example as a part of a
larger framework built around 6D rotation representation, the
regularization loss imposed by the prior is just the negated log
probability density of the pose vector.

When optimizing in the latent space, which is, in general,
preferred for inverse kinematics tasks, care must be taken to
compute the loss as the negated log probability density of the
projection into the ambient space (rather than just the density
of the multivariate normal) due to input-dependent change of
volume of the transformation. To make latent space optimization
in presence of additional constraints imposed on the 6D rotation
representation more efficient, we extended the forward pass of
RealNVP (that is, from the latent space to the ambient space)
such that the log probability density of the ambient space
projection is computed along with the projection itself. This
reduces three-fold the number of passes through RealNVP.

\section{Empirical Evaluation}
\label{sec:evaluation}

Our neural prior is trained on a dataset consisting of $\approx$
1.4 million poses using Adam optimizer with early stopping on
a held-out validation set. In the experiments below, we compare
the trained prior with the data distribution, with the VPoser
prior trained \textit{on the same} dataset, as well as explore
the influence of the inverse Gram-Schmidt augmentation on the
learned prior distribution.

\subsection{Quantitative Performance Metrics}

\begin{figure}
\centering
\includegraphics[width=0.75\linewidth]{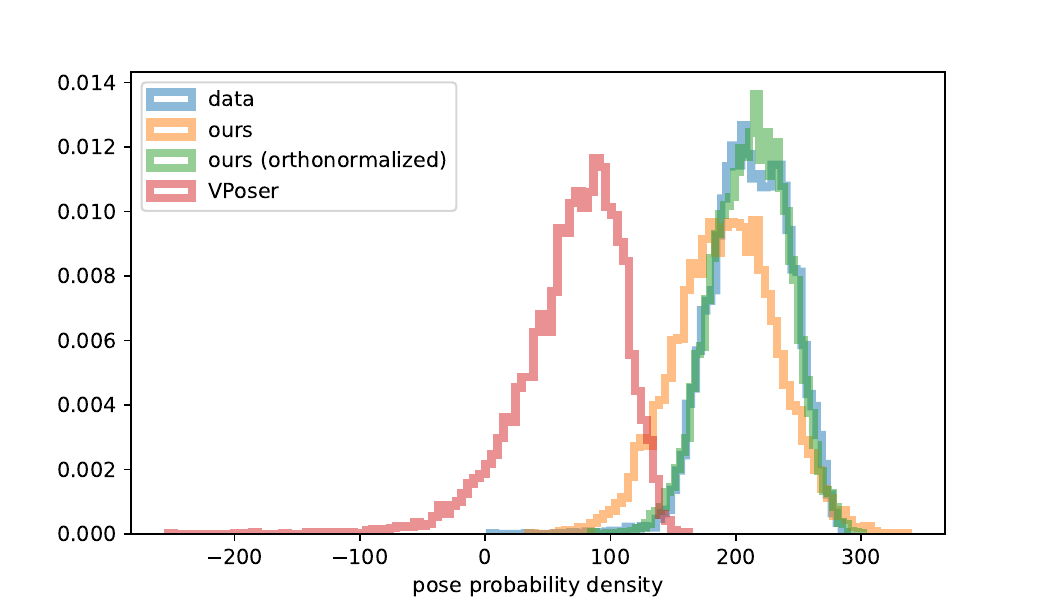}
\begin{tabular}{r | c c c}
\toprule
 & Ours raw & Ours orthonormalized & VPoser \\
\midrule
 KS statistics & 0.24 & 0.020 & 0.98 \\
\bottomrule
\end{tabular}
\caption{Data vs. prior distribtions. Our prior is much closer
to the data than VPoser's.}
\label{fig:ks}
\end{figure}

To quantitatively evaluate the quality of the learned prior, we
compare, through sampling, the learned prior distribution to the
data distribution. As a baseline, we perform a similar
comparison with the VPoser-induced prior distribution. We
generate $10\,000$ samples from the prior, draw randomly the
same number of entries from the data set, and compute the log
probability density of each rotation vector. We then compare
the \textit{distribution of densities}. If the sample
distribution is close to the data distribution the distributions
of densities will be close as well. Conversely, if the
distributions of densities are far away from each other, the
learned prior distribution does not approximate the data
distribution well.

The distributions of densities are shown in Figure~\ref{fig:ks}.
For our prior, we show two distributions: of raw and of
orthonormalized samples, since 6D rotations in the data are
orthonormal. The samples from VPoser are genuinely in the three-dimensional
rotation vector representation, and we convert them into orthonormal
6D rotations. One can see that our prior's learned distribution
is close to the data distribution, and the distribution of
densities of orthonormalized samples is virtually
indistinguishable from that of the data. On the contrary,
VPoser produces samples with much lower density on average,
and apparently the VPoser's induced prior differs a lot from the
data distribution. 

To quantify the results, we compute the Kolmogorov-Smirnov (KS)
statistics between the densities of the samples from the priors
and the data. The statistics belongs to the range $[0, 1]$; the
smaller the statistics is, the closer the distributions. For
orthonormalized samples from our prior the KS statistic is
0.02 (very close). For VPoser, the statistic is 0.98 (barely
overlapping).

\subsection{Qualitative Assessment}

\begin{figure}
    \centering
    \begin{subfigure}[c]{0.44\linewidth}
        \centering
        \includegraphics[width=\linewidth]{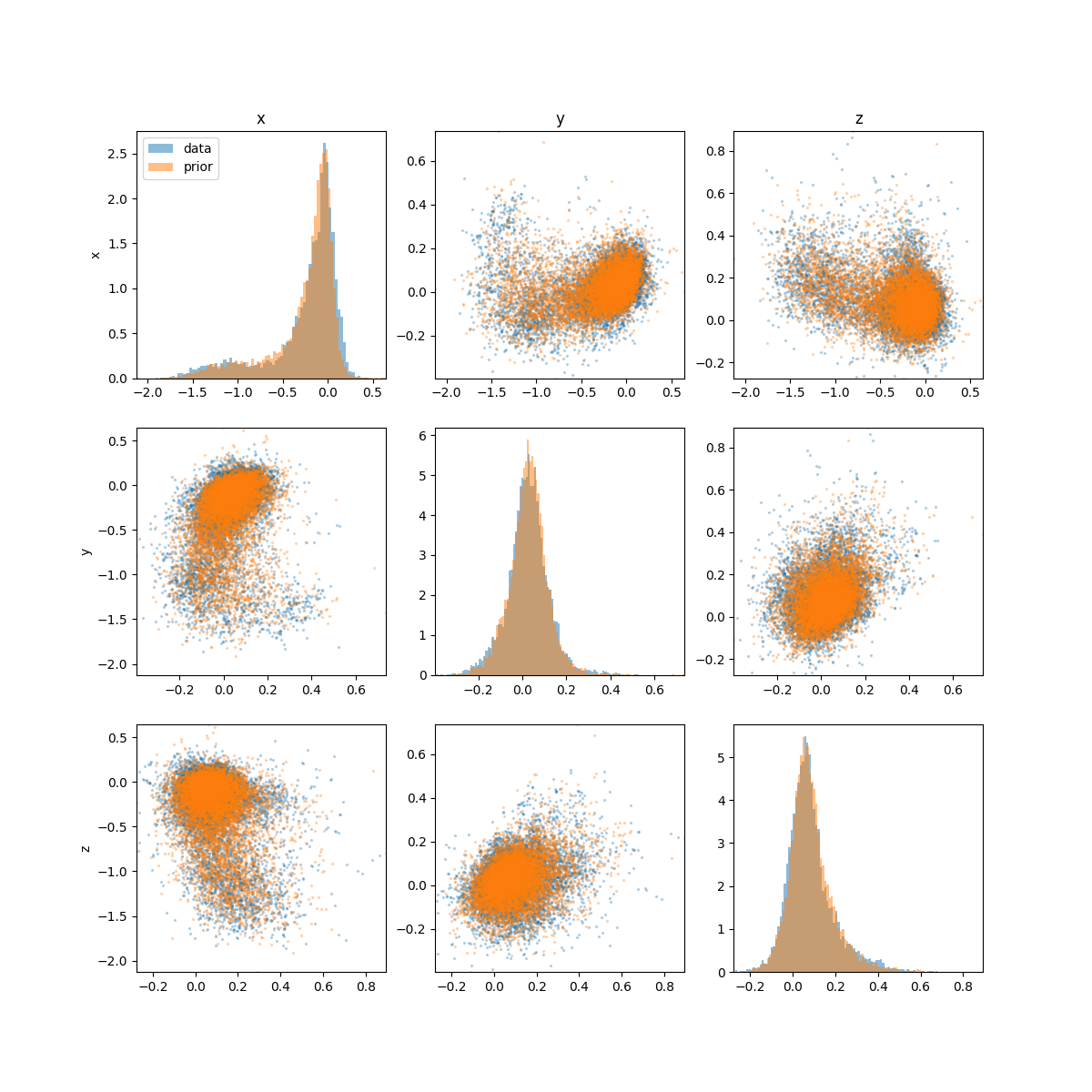}
        \caption{left hip, ours}
        \label{fig:left-hip}
    \end{subfigure}
    \begin{subfigure}[c]{0.44\linewidth}
        \centering
        \includegraphics[width=\linewidth]{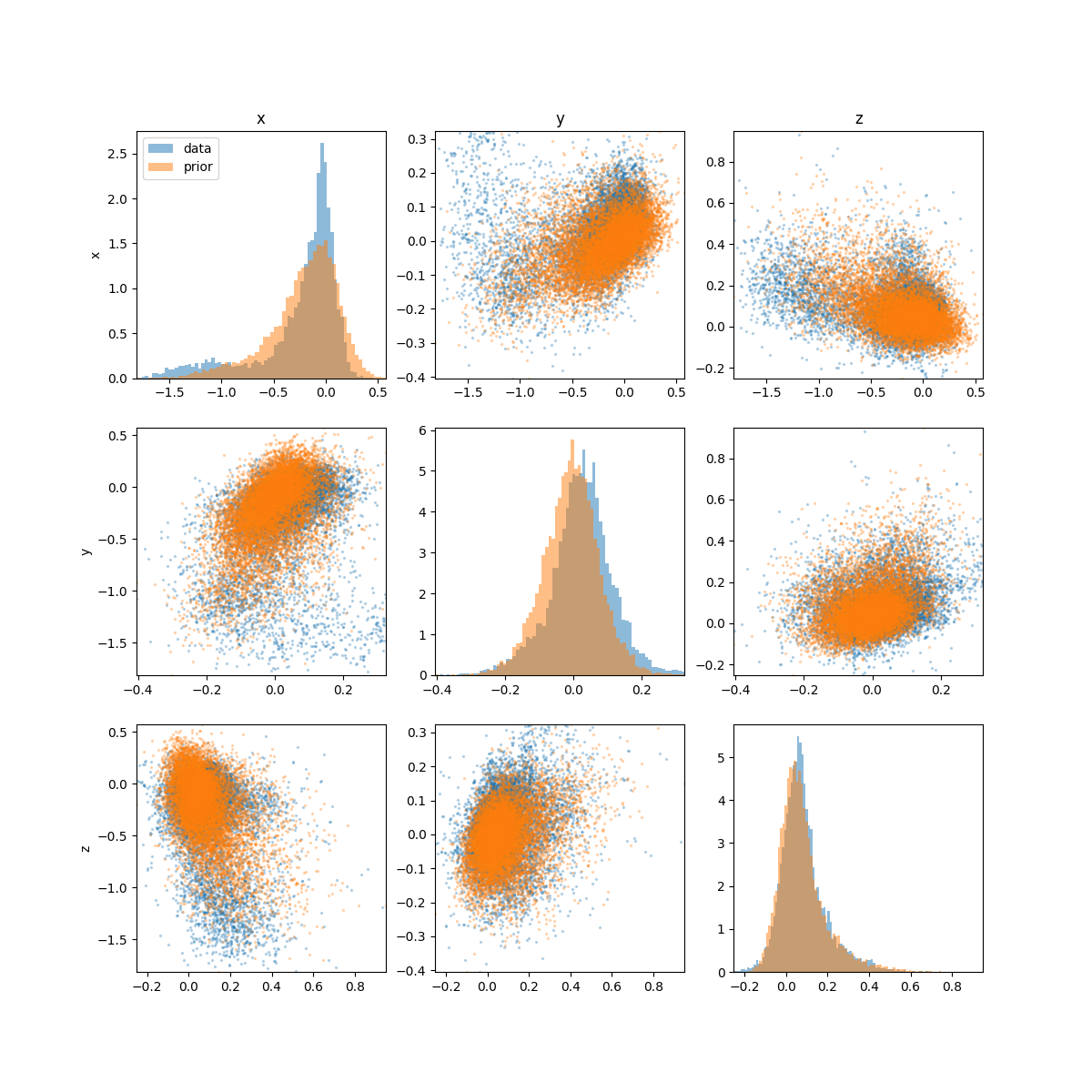}
        \caption{left hip, VPoser}
        \label{fig:vposer-left-hip}
    \end{subfigure}
    \begin{subfigure}[c]{0.44\linewidth}
        \centering
        \includegraphics[width=\linewidth]{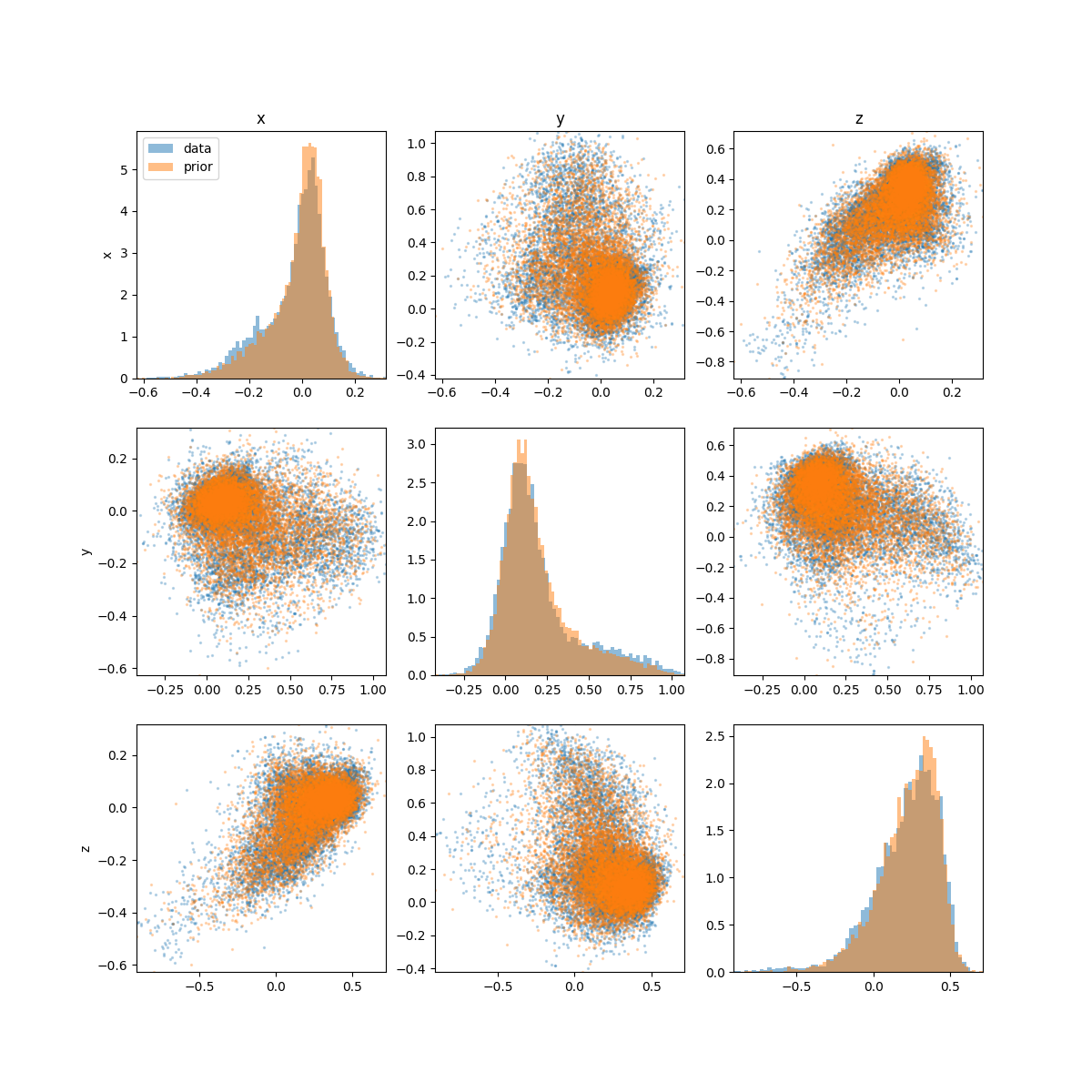}
        \caption{right collar, ours}
        \label{fig:right-collar}
    \end{subfigure}
    \begin{subfigure}[c]{0.44\linewidth}
        \centering
        \includegraphics[width=\linewidth]{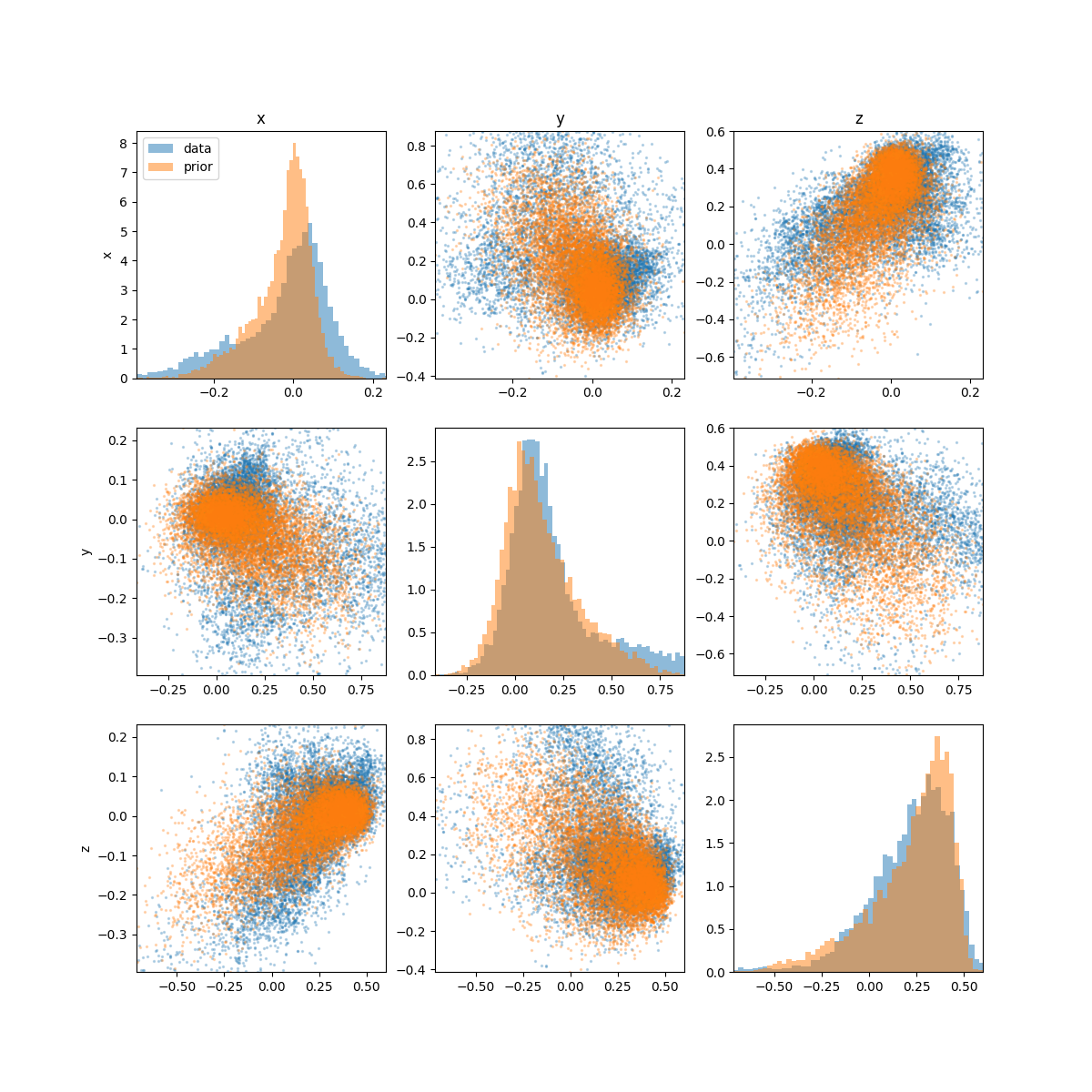}
        \caption{right collar, VPoser}
        \label{fig:vposer-right-collar}
    \end{subfigure}
    \begin{subfigure}[c]{0.44\linewidth}
        \centering
        \includegraphics[width=\linewidth]{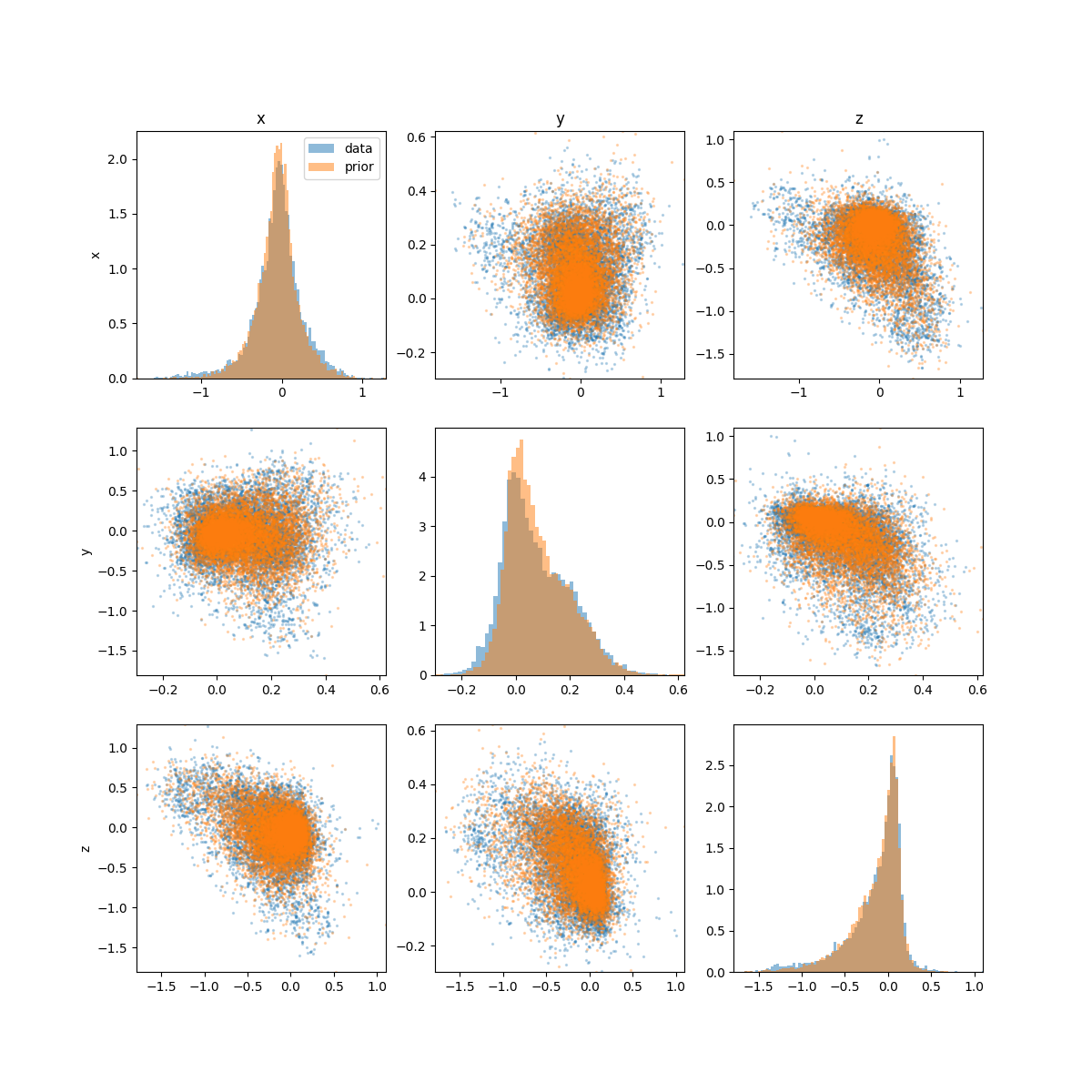}
        \caption{right wrist, ours}
        \label{fig:right-wrist}
    \end{subfigure}
    \begin{subfigure}[c]{0.44\linewidth}
        \centering
        \includegraphics[width=\linewidth]{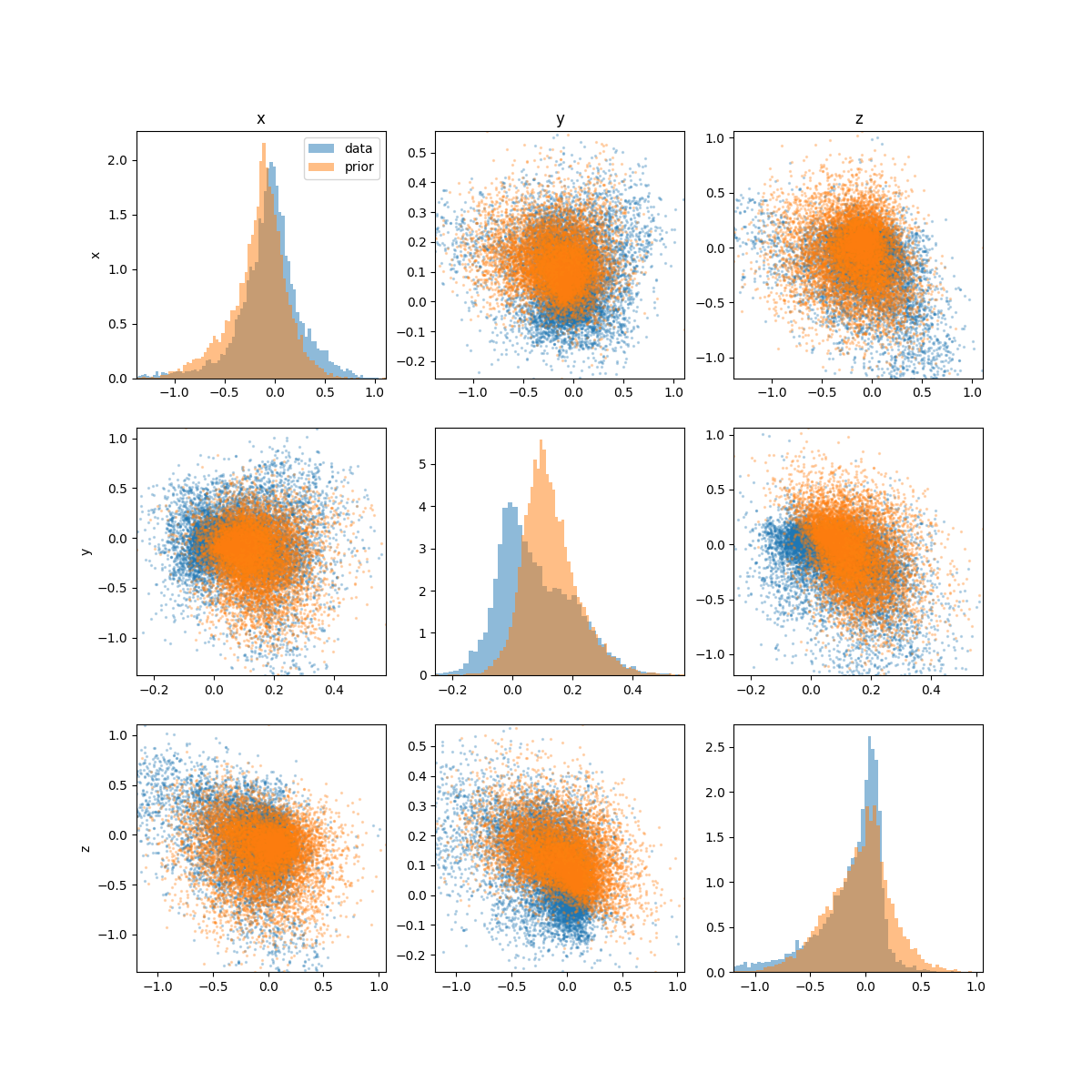}
        \caption{right wrist, VPoser}
        \label{fig:vposer-right-wrist}
    \end{subfigure}
    \caption{Marginal data and prior distributions of some joints,
    projected to the rotation vector space.  Our prior aligns with the data
    distribution almost perfectly. There are noticeable discrepancies between
    the data distribution and the VPoser prior trained on the data.}
    \label{fig:joint-marginals}
\end{figure}

It is common for qualitative assessment of the pose prior to render
sample poses; however, such visualization can be deceiving: on one hand,
the author is tempted to hand-pick poses for display, on the other hand,
the reader and the author may not always agree whether the presented
poses are natural yet diverse. Thus, instead of selected sampled poses, we 
show marginal data and learned prior distributions of a few joints of
our body model's skeleton (Figure~\ref{fig:joint-marginals}). For 
better visualization as matrix plots, we convert the pose vector from 6D 
rotations to 3D rotation vectors.

We compare the marginal distribution for three joints in different parts of the skeleton and at the different distances from the root joint: the left hip, the right collar bone, and the right wrist. The left-hand plots show samples from our prior, and the right-hand plots --- from the VPoser prior (orange), superimposed with the samples from the data set (blue). The samples from our prior align almost  perfectly with the data. There are noticeable discrepancies between the data samples and the samples drawn from the VPoser prior. 

\subsection{Ablation Study}

\begin{figure}
    \centering
    \begin{subfigure}[c]{0.49\linewidth}
        \centering
        \includegraphics[width=\linewidth]{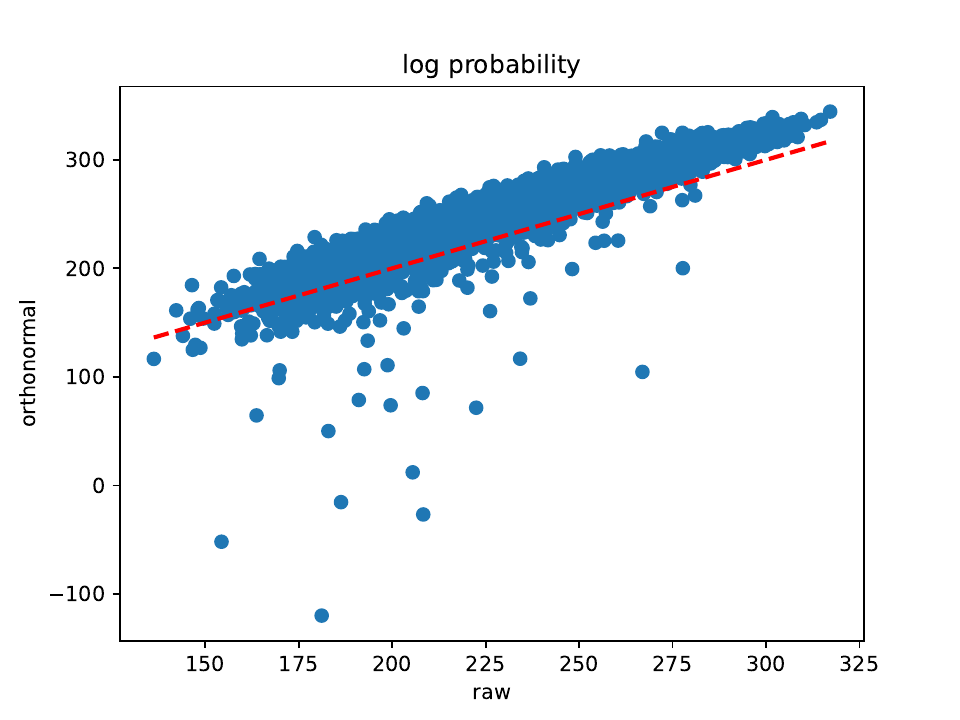}
        \caption{trained on raw data}
        \label{fig:no-dgs}
    \end{subfigure}
    \begin{subfigure}[c]{0.49\linewidth}
        \centering
        \includegraphics[width=\linewidth]{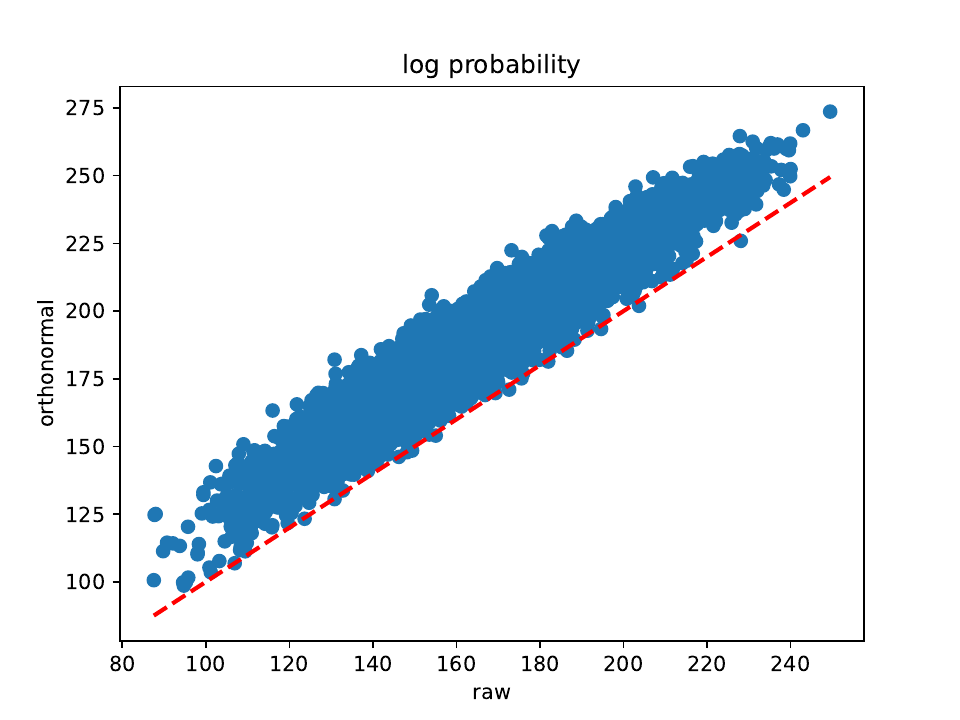}
        \caption{trained on data augmented through inverse Gram-Schmidt}
        \label{fig:dgs}
    \end{subfigure}
    \caption{Densities of orthonormalized vs. raw samples from
    the prior. Inverse Gram-Schmidt helps prevent
    discontinuities in the latent space and results in a
    smoother model.}
    \label{fig:dgs-no-dgs}
\end{figure}

In Section~\ref{sec:degramschmidt} we introduced data
augmentation through the inverse Gram-Schmidt process.
In this section, we subject this augmentation
to an ablation study to show that inverse Gram-Schmidt 
indeed improves the model and is crucial for the model's 
robustness. Figure~\ref{fig:dgs-no-dgs} compares two models. 
The left-hand model (Figure~\ref{fig:no-dgs}) was trained
on raw data, the right-hand model (Figure~\ref{fig:dgs}) was 
trained on data augmented by inverse Gram-Schmidt. Both models, 
by design or due to inherent imperfections of learned transformation
between the latent and the ambient spaces, produce samples which are
not necessarily orthonormal. The horizontal axis measures the density
of the raw samples, the vertical axis --- the density of the
corresponding orthonormalized pose vectors. The dashed red 
lines pass through the equal densities of raw and
orthonormalized samples.

When the model is trained on raw data, orthonormalized samples occasionally 
have a much lower density than their raw counterparts 
(points below the red line in Figure~\ref{fig:no-dgs}). This likely happens
because the model struggled with learning the density of a lower-dimensional
manifold within the 6D rotation vector space.  However, when the model is
trained on the inverse Gram-Schmidt augmented data, 
the density of an orthonormalized sample is virtually always 
higher than that of the corresponding raw sample, as anticipated (almost all points are on or above the red line in Figure~\ref{fig:dgs}).

\section{Conclusion}
\label{sec:conclusion}

We have presented a systematic and theoretically grounded
approach to constructing a neural prior over human body poses
using normalizing flows. Our formulation addresses key
challenges in pose representation, such as the incompatibility
of 6D rotation vectors with standard density estimation
frameworks, by introducing an inverse Gram-Schmidt augmentation
scheme. The resulting model is expressive, stable to train, and
efficient to use during inference, supporting both ambient and
latent space optimization.

This neural pose prior integrates seamlessly into motion capture
pipelines and enables principled regularization in
optimization-based tasks. Our design choices are justified
through both qualitative and quantitative evaluation, as well as
ablation studies that highlight the impact of key components.

A promising direction for future work is extending the pose
prior to a body shape-conditioned model, where the pose
distribution is adapted dynamically to individual body shapes.
Notably, normalizing flows in general and the RealNVP
architecture in particular support conditioning in
principle~\cite{PPM17}. While some architectural modifications and tuning
will be required to ensure optimal performance, this extension
could yield personalized priors that further improve anatomical
plausibility and motion realism.

\bibliographystyle{acm}
\bibliography{refs}

\end{document}